\documentclass[10pt,twocolumn,letterpaper]{article}

\usepackage{cvpr}              

\newcommand{\methodname}{{{VLG}}\xspace}

\newcommand{\figref}[1]{Fig.~\ref{#1}}
\newcommand{\secref}[1]{Sec.~\ref{#1}}

\newcommand{\tabref}[1]{Table~\ref{#1}}

\usepackage{booktabs}

\definecolor{myblue}{HTML}{A4B7DC}
\definecolor{myorange}{HTML}{CC915A}
\definecolor{mygreen}{HTML}{97CE9B}
\definecolor{myred}{HTML}{B25B58}

\definecolor{cvprblue}{rgb}{0.21,0.49,0.74}
\usepackage[pagebackref,breaklinks,colorlinks,allcolors=cvprblue]{hyperref}

\def\paperID{*****} 
\def\confName{CVPR}
\def\confYear{2025}

\title{Towards Vision-Language-Garment Models\\ \small{For Web Knowledge Garment Understanding and Generation}}

\author{Jan Ackermann~~~George Nakayama~~~Guandao Yang~~~Tong Wu~~~Gordon Wetzstein\\
\\[-0.4cm]
Stanford University
}

\begin{document}
\maketitle
\begin{abstract} Multimodal foundation models have demonstrated strong generalization, yet their ability to transfer knowledge to specialized domains such as garment generation remains underexplored. We introduce \methodname, a vision-language-garment model that synthesizes garments from textual descriptions and visual imagery. Our experiments assess \methodname's zero-shot generalization, investigating its ability to transfer web-scale reasoning to unseen garment styles and prompts. Preliminary results indicate promising transfer capabilities, highlighting the potential for multimodal foundation models to adapt effectively to specialized domains like fashion design. The project page is \url{https://www.computationalimaging.org/publications/vision-language-garment-models/}.\end{abstract}    
\vspace{-0.5cm}
\section{Introduction}
\label{sec:intro}
Multimodal foundation models have demonstrated remarkable generalization and reasoning capabilities across various tasks, driven by extensive pre-training on web-scale datasets. However, their effectiveness in transferring these general-purpose abilities to specialized domains—such as garment design and generation—remains relatively unexplored. Garment design presents a challenging testbed for these models due to its inherent multimodal nature, requiring the integration of visual aesthetics, textual descriptions, functional constraints, and user-specific preferences.

Traditional garment generation approaches have predominantly relied on retrieval-based or narrowly supervised learning methods~\cite{liu2023sewformer, he2024dresscodeautoregressivelysewinggenerating}, which often struggle to generalize beyond their training data and exhibit limited flexibility in addressing complex, nuanced user inputs. While recent advancements leveraging large-scale datasets like GarmentCode~\cite{korosteleva2023garmentcode} have improved multimodal garment generation~\cite{bian2024chatgarment, nakayama2024aipparel}, these methods do not explore model-performance out of their data training distribution.

In this work, we explore the potential of vision-language foundation models to transfer their multimodal reasoning skills into garment generation tasks effectively. Specifically, we introduce and evaluate \methodname, a Vision-Language-Garment model built upon a pre-trained vision-language foundation model. Our core contributions are as follows:

\begin{enumerate}
    \item The introduction of \methodname, a multimodal garment generation model designed to leverage generalized vision-language reasoning for garment-specific tasks.
    \item An evaluation of the reasoning capabilities of \methodname against state-of-the-art methods, using a curated benchmark to assess multimodal knowledge transfer.
\end{enumerate}

Our study aims to illuminate how multimodal foundation models can adapt their generalized reasoning abilities to specialized, domain-specific applications, laying the groundwork for future research in multimodal knowledge transfer and intelligent garment generation.

\begin{figure}[t] \centering \includegraphics[width=\linewidth]{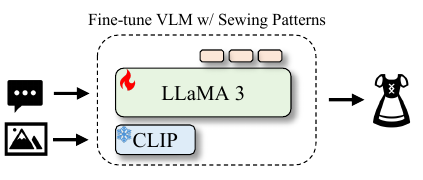} \caption{\textbf{Vision-Language-Garment Models.} We tune existing vision-language models for garment-specific tasks to predict sewing patterns from multimodal (textual and visual) inputs.} \label{fig:teaser} \end{figure}
\begin{figure*}[t] \centering
 \includegraphics[width=\linewidth]{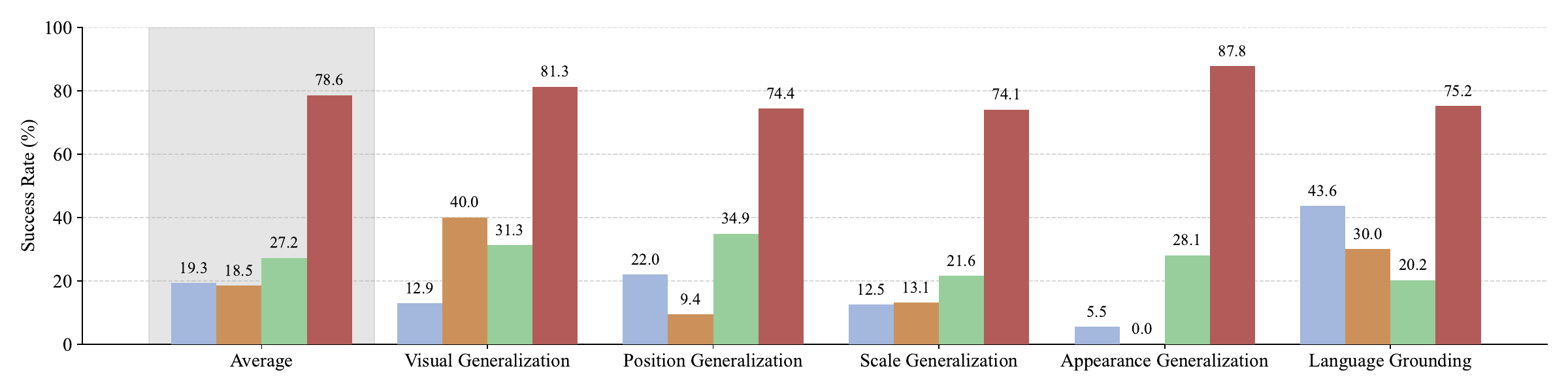}
 \caption{\textbf{Reasoning Transfer Evaluation.} Performance comparison of our proposed model~(\textcolor{myred}{red}) against DressCode~\cite{he2024dresscodeautoregressivelysewinggenerating}~(\textcolor{myblue}{blue}), SewFormer~\cite{liu2023sewformer}~(\textcolor{myorange}{orange}) and Sewformer-FT~(\textcolor{mygreen}{green}), across various reasoning dimensions.} \label{fig:main}
\end{figure*}
 
\section{Related Work} \label{sec:related}
\paragraph{Digital Garment Understanding and Generation.}
Advances in digital garment generation have explored a sewing pattern representation~\cite{liu2023sewformer, NeuralTailor2022, he2024dresscodeautoregressivelysewinggenerating} that stands out due to its direct application to real-world manufacturing and garment production processes. The recent availability of larger-scale sewing pattern datasets like GarmentCode~\cite{korosteleva2023garmentcode} has facilitated multimodal learning approaches~\cite{bian2024chatgarment, nakayama2024aipparel}, significantly enhancing flexibility and cross-modal understanding. Nonetheless, there is no clarity on the pre-trained backbone's role in the final model performance.
\paragraph{Multimodal Foundation Models and Knowledge Transfer.} Multimodal foundation models like GPT-4~\cite{openai2024gpt4technicalreport} and LLaMA~\cite{touvron2023llama2openfoundation} have demonstrated strong generalization capabilities across diverse tasks. Recent works explore adapting these models to specialized domains through prompting~\cite{hugginggpt, wu2023visualchatgpttalkingdrawing} and domain-specific fine-tuning~\cite{liu2023llava, zhu2023minigpt, tong2024cambrian1}, though fine-tuning often risks catastrophic forgetting. Notably, robotics-focused approaches such as RT-2~\cite{rt22023arxiv} and OpenVLA~\cite{kim24openvla} successfully retain general reasoning during transfer. Motivated by these results, our work investigates how effectively multimodal reasoning transfers to the garment domain by evaluating adapted models on a garment-specific reasoning benchmark.
\section{Vision-Language-Garment Models} \label{sec:method}
This work explores the capability of multimodal foundation models to transfer their general knowledge and reasoning skills into the specialized domain of garment generation. 
In the following sections, we detail our approach, including dataset curation for training and evaluating reasoning transfer, and the fine-tuning procedure to adapt pre-trained multimodal foundation models to the garment domain.
\subsection{Dataset Curation} \label{sec:dataset}
A critical factor for multimodal reasoning transfer to garment-specific tasks is the curation of suitable training and testing datasets. We contribute two key datasets designed explicitly to train and benchmark multimodal garment reasoning abilities.
\paragraph{Garment Prompt Dataset.} Existing garment datasets such as AIpparel~\cite{nakayama2024aipparel} and DressCode~\cite{he2024dresscodeautoregressivelysewinggenerating} predominantly feature detailed and precise textual descriptions, differing significantly from naturally vague and emotionally guided human interactions. To bridge this gap, we leverage GPT-4o's natural language generation abilities to create a dataset with diverse, context-rich, and realistically ambiguous prompts aligned closely with human interaction patterns. Given sewing pattern images from the GarmentCode dataset, GPT-4o was prompted to generate diverse and nuanced queries, expanding the coverage of naturalistic prompts significantly beyond existing benchmarks.
\paragraph{Synthetic Textured Garment Dataset.} Evaluating garment reasoning requires data accurately reflecting real-world variability in garment appearances and textures. Previous approaches often rely on manually designed textures, limiting scalability and generalization. We introduce a scalable pipeline to address this and generate diverse, textured garment images. Our method begins with rendered garment images and utilizes the Canny Edge Detector to derive control masks. GPT-4o~\cite{openai2024gpt4technicalreport} is then queried to generate plausible textual descriptions for textures, which are subsequently synthesized using ControlNet~\cite{zhang2023adding} into neural-rendered images. The backgrounds are masked white to maintain consistency and ensure clean visual boundaries.
For evaluation, we refine some synthesized images using a Virtual Try-On tool~\cite{choi2024improving}. Given the limitations of such tools in geometry preservation, we manually curate the final dataset, ensuring suitability for rigorous multimodal reasoning evaluation.

\subsection{Model Training and Fine-Tuning} \label{sec:pretrain}
For the training, we utilize AIpparel's~\cite{nakayama2024aipparel} garment tokenization scheme to represent garments.
We adopt standard fine-tuning practices of multimodal foundation models. Specifically, we train 16 Nvidia H100 GPUs with a per-GPU batch size of 6 and gradient accumulation steps of 5, achieving an effective batch size of 480. Training proceeds for 20 epochs, employing a linear warm-up of the learning rate from $0$ to $6\cdot 10^{-5}$ during the first $1000$ steps, followed by linear decay to $0$.
To maintain the base multimodal model's general reasoning capabilities and visual knowledge, we freeze the vision encoder and cross-attention modules. The final layer of the regression head is initialized to zero, stabilizing early-stage optimization by controlling the magnitude of prediction errors.
\section{Experiments}
\label{sec:experiments}
In this section, we present qualitative and quantitative results of our method, comparing against the following baselines: DressCode~\cite{he2024dresscodeautoregressivelysewinggenerating}, SewFormer~\cite{liu2023sewformer}, and SewFormer-FT, a SewFormer fine-tuned on GarmentCodeData~\cite{korosteleva2024garmentcodedatadataset3dmadetomeasure}. When an input data-point has mismatched modality, we employ GPT-4o to translate it into the correct modality.

\paragraph{Metrics.} Throughout this section, we quantitavely evaluate the garment predictions by garment accuracy~\cite{nakayama2024aipparel} and Vertex L2. For text-based predictions, we ask human evaluators to score the overall-quality and the prompt-alignment.

\paragraph{Garment Reasoning}
Analogous to VLA approaches~\cite{rt22023arxiv, kim24openvla}, we analyze how existing reasoning capabilities transfer to the fine-tuned model. To evaluate this, we created a synthetic dataset applying variations in backgrounds (visual generalization), translation (position generalization), zooms (scale generalization), and random textures (appearance generalization). For language grounding, we utilize a held-out text test set.

The results are summarized in \figref{fig:main}. Models without pre-training struggle with out-of-distribution (OOD) data, whereas our \methodname benefits significantly from robust pre-training. Additionally, baselines perform particularly well on language grounding, benefiting from GPT-4o's inherent reasoning capabilities.

\begin{figure}[th]
    \centering
    \includegraphics[width=\linewidth]{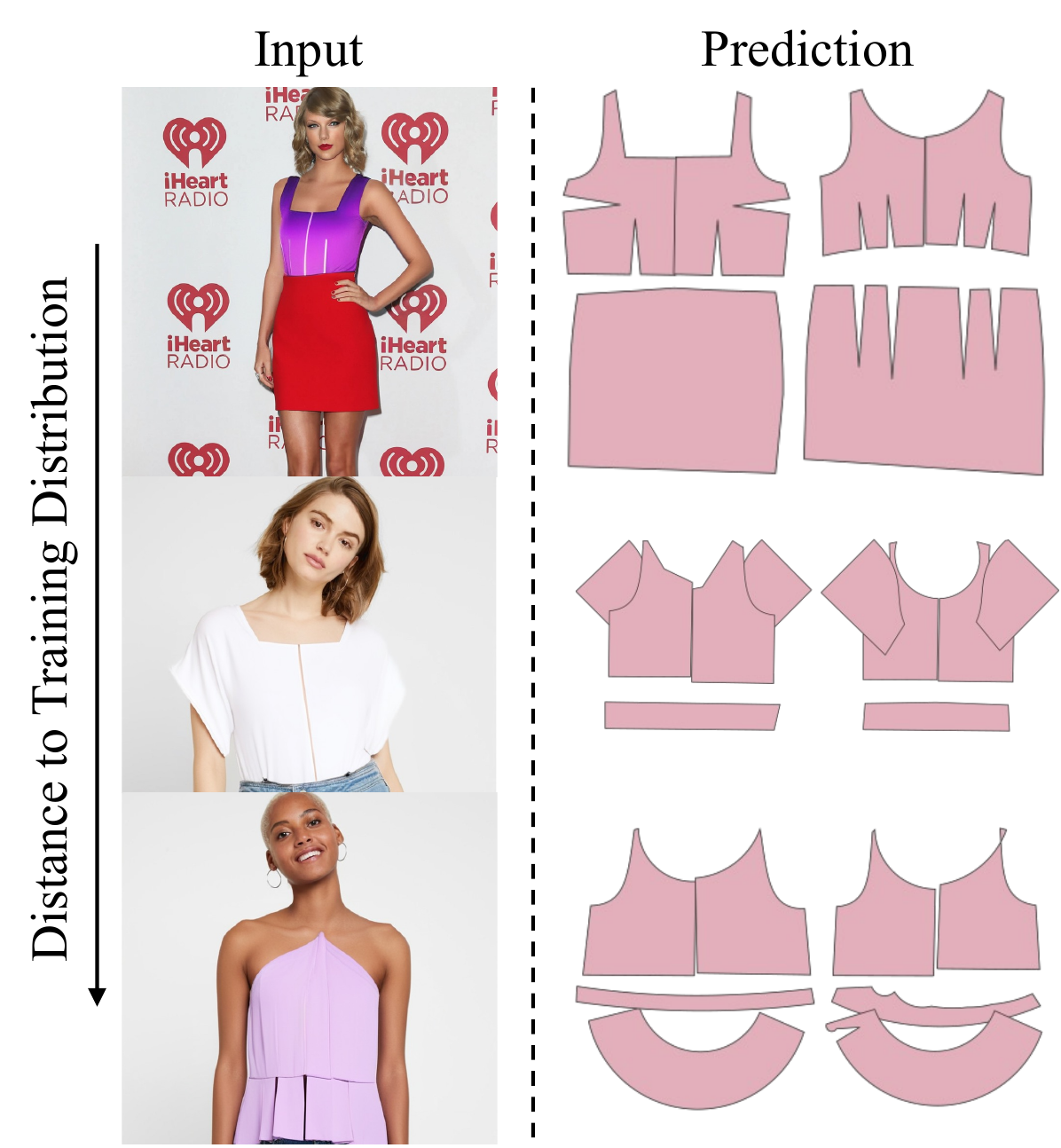}
    \caption{\textbf{Visual Generalization.} Pattern predictions degrade as input becomes increasingly distant from the training distribution.}
    \label{fig:vis-generalization}
\end{figure}

\begin{figure}[th]
    \centering
    \includegraphics[width=\linewidth]{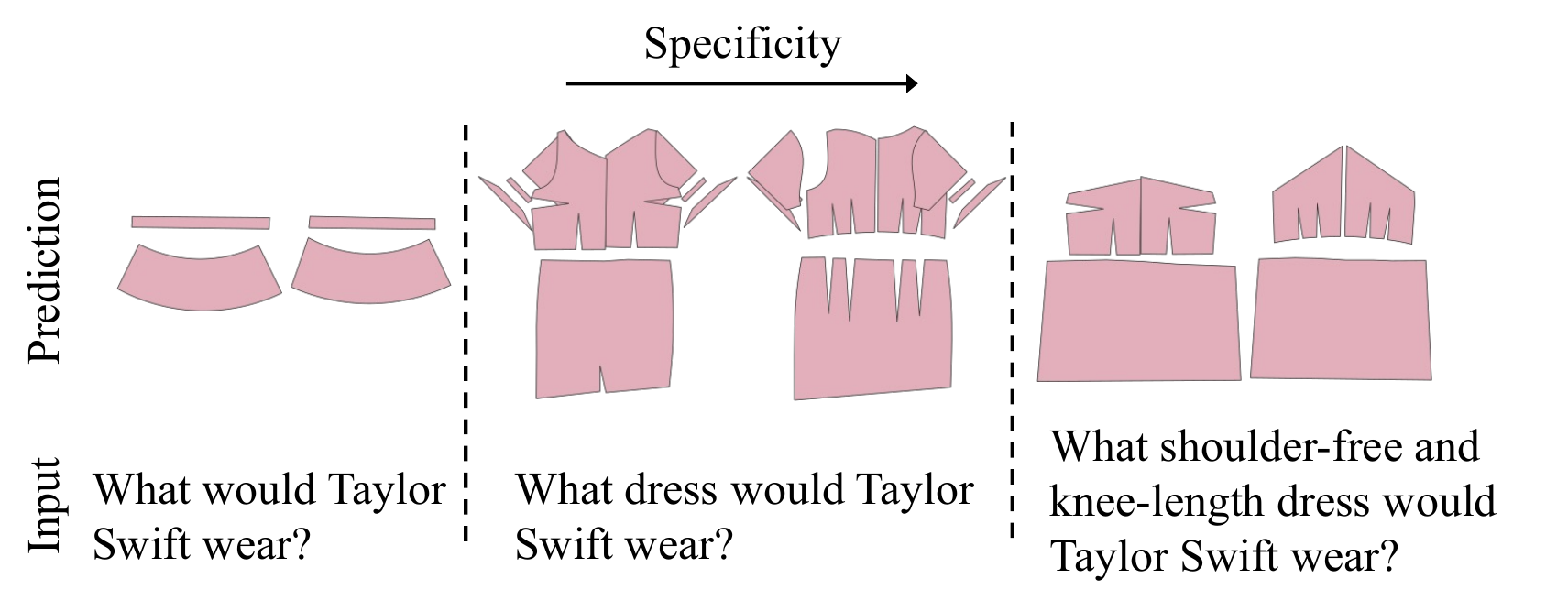}
    \caption{\textbf{Textual Generalization.}  Increasing specificity in textual prompts leads to more accurate garment pattern predictions.}
    \label{fig:text-generalization}
\end{figure}

\paragraph{Limits of Reasoning}
We further investigate how well the model generalizes across novel images and texts. We evaluate predictions on images discussed in \secref{sec:dataset}. As shown in \figref{fig:vis-generalization}, our model predicts sewing patterns from unseen real-world images, with quality dependent on similarity to the training data. The closest input yields the best predictions, while the most dissimilar input yields poor results.

Additionally, we test model performance on OOD texts (\figref{fig:text-generalization}), finding that the model struggles with under-specified prompts but significantly improves when familiar words, even in new contexts, are included.

These results indicate that the model effectively learns the training data distribution and highlight the importance of an extensive and varied dataset.

\begin{table}[th]
\centering
\begin{tabular}{lccc}
\toprule
\textbf{Method} & \textbf{Garment Acc.} $\uparrow$ & \textbf{Vertex L2} $\downarrow$ & \textbf{Text Acc.} $\uparrow$ \\
\midrule
Base & \textbf{0.755} & \textbf{4.896} & 0.388 \\
+Prompts & 0.727 & 5.483 & \textbf{0.682} \\
+Textures & 0.772 & 5.047 & 0.666 \\
\bottomrule
\end{tabular}
\caption{\textbf{Dataset Ablation.} Comparison of methods across Garment Accuracy, Vertex L2, and Text Accuracy.}
\label{tab:metrics}
\end{table}
\vspace{-0.8cm}
\begin{figure}[th]
    \centering
    \includegraphics[width=\linewidth]{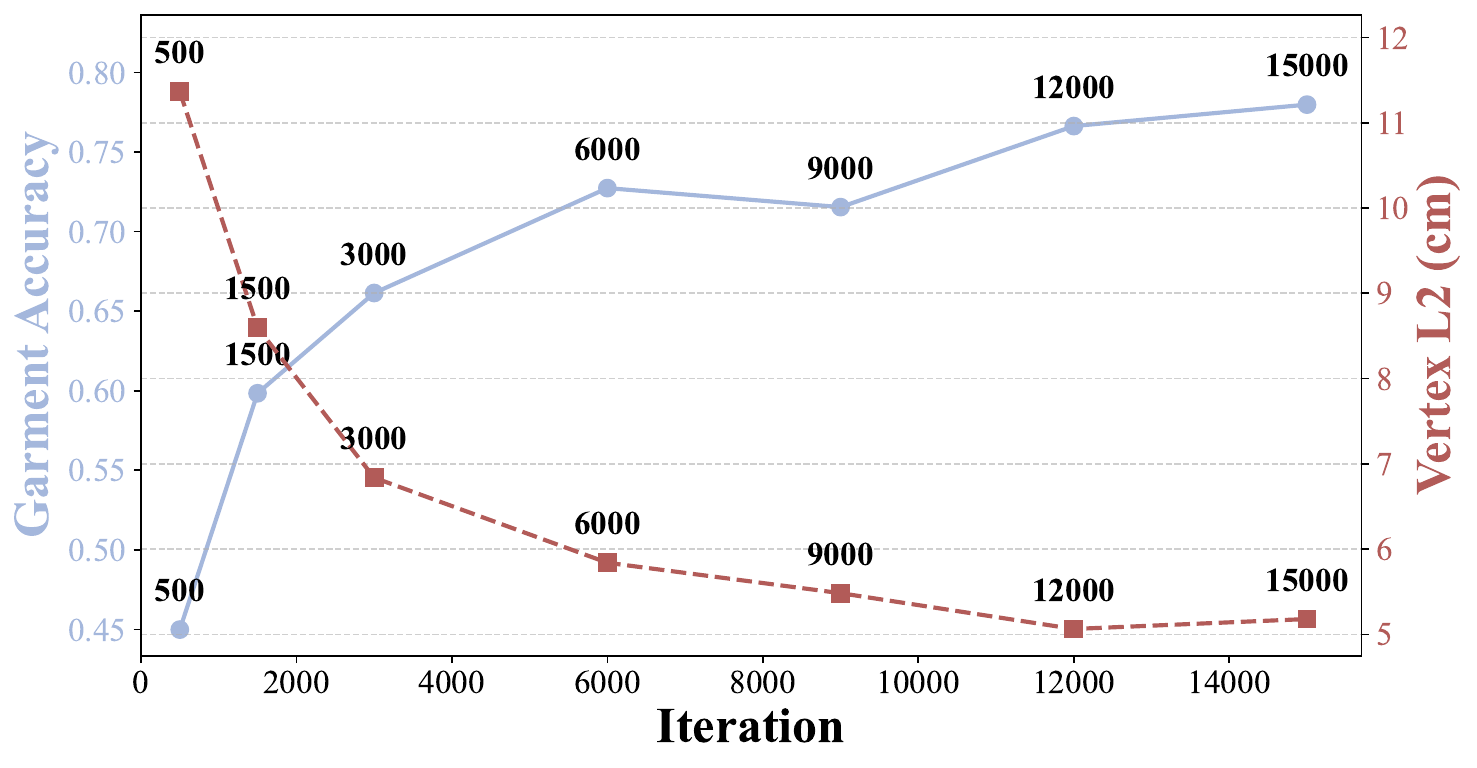}
    \caption{\textbf{Validation Loss by Iteration.} Garment accuracy steadily improves, and Vertex L2 loss decreases over training iterations, indicating successful model convergence.}
    \label{fig:iterations}
\end{figure}

\begin{figure}[th]
    \centering
    \includegraphics[width=\linewidth]{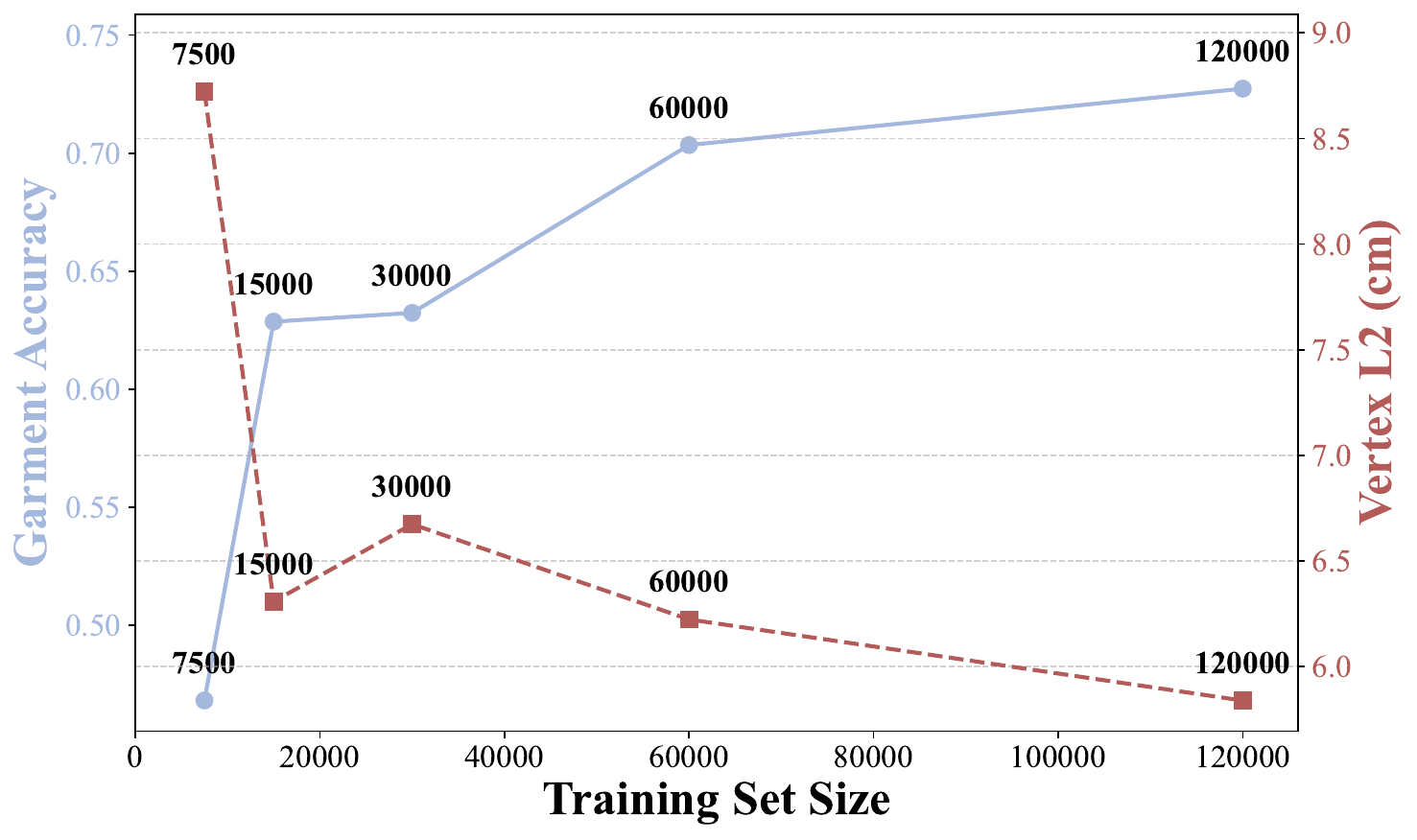}
    \caption{\textbf{Validation Loss by Data Size.} Increasing the training dataset size leads to improved garment accuracy and reduced vertex prediction error, highlighting the importance of dataset scale.}
    \label{fig:data-size}
    \vspace{-0.5cm}
\end{figure}
\vspace{-1cm}
\paragraph{Ablations}
Given that we fine-tune a vision-language model on a novel domain with newly introduced tokens, we analyze its convergence behavior. 

As shown in \tabref{tab:metrics}, augmenting the base model with varied text data notably improves text accuracy with minimal reduction in image-based metrics. However, adding textured garments offers minimal improvement in visual generalization. All models were trained with identical compute budgets; thus, introducing new data types slightly reduces the number of iterations per data type.

\figref{fig:iterations} illustrates the training and validation loss curves, both demonstrating gradual convergence over time.  

\figref{fig:data-size} further examines validation loss after training for 10 epochs with varying training data sizes, consistently using a validation set of 5000 garments. A clear trend emerges, indicating that larger datasets improve both garment accuracy and vertex L2 metrics.

\vspace{-0.2cm}
\section{Discussion}
\label{sec:conclusion}
\vspace{-0.2cm}
\paragraph{Limitations.}
Our experiments confirm that vision-language garment models (\methodname) benefit significantly from fully data-driven approaches and strong vision-language pre-training. However, the limited inductive biases imply challenges in generalizing to out-of-distribution scenarios, as evidenced by performance drops in novel and underspecified inputs. Future research should prioritize expanding dataset size and diversity to overcome these limitations.

\vspace{-0.5cm}
\paragraph{Conclusion.}
We introduced \methodname, demonstrating their effectiveness in developing garment-focused foundation models. The results show that \methodname scales efficiently with increased data and leverages the strengths of pre-trained vision-language models, underlining their potential for future advancements in garment modeling.

\vspace{-0.3cm}
\section{Acknowledgements}
\label{sec:ack}
\vspace{-0.2cm}
We thank LVMH and Google for their support. We also thank Stanford's Marlowe cluster for providing GPU computing resources for model training and evaluation.

{
    \small
    \bibliographystyle{ieeenat_fullname}
    \bibliography{main}
}


\end{document}